\documentclass[10pt,conference]{IEEEtran}

\usepackage{cite}

\ifCLASSINFOpdf
   \usepackage[pdftex]{graphicx}
   \usepackage{xcolor}
   \graphicspath{{figs/}}
   \DeclareGraphicsExtensions{.pdf,.jpeg,.png}
\else
   \usepackage[dvips]{graphicx}
   \graphicspath{{../figs/}}
   \DeclareGraphicsExtensions{.eps}
\fi
\usepackage[cmex10]{amsmath}
\usepackage{amsthm}

\usepackage{algorithmic}

\usepackage{array}

\ifCLASSOPTIONcompsoc
  \usepackage[caption=false,font=normalsize,labelfont=sf,textfont=sf]{subfig}
\else
  \usepackage[caption=false,font=footnotesize]{subfig}
\fi
\usepackage{url}

\begin{document}
%
\title{Cross-Domain Deep Face Matching for Real Banking Security Systems}


\author{\IEEEauthorblockN{
Johnatan S. Oliveira$^{1,*}$, Gustavo B. Souza$^{2,*}$, Anderson R. Rocha$^{3}$, Fĺavio E. Deus$^{1}$ and Aparecido N. Marana$^{4}$
}
\IEEEauthorblockA{
$^1$ Department of Electrical Engineering, University of Bras\'{i}lia (UnB), Bras\'{i}lia, Brazil.\\
$^2$ Department of Computing, Federal University of S\~{a}o Carlos (UFSCar), S\~{a}o Carlos, Brazil.\\
$^3$ Institute of Computing, University of Campinas (Unicamp), Campinas, Brazil.\\
$^4$ Department of Computing, S\~{a}o Paulo State University (Unesp), Bauru, Brazil.\\
$^{*}$ Equal contributors.
\\E-mails: \{jow, gustavo.botelho\}@gmail.com, anderson.rocha@ic.unicamp.br,\\
flavioelias@unb.br and nilceu@fc.unesp.br\\ \\
\textbf{--- PAPER ACCEPTED AT  ICEDEG 2020 (PLEASE CITE ITS PROCEEDINGS) ---}
}
}

\maketitle

\begin{abstract}
Ensuring the security of transactions is currently one of the major challenges that banking systems deal with. The usage of face for biometric authentication of users is attracting large investments from banks worldwide due to its convenience and acceptability by people, especially in cross-domain scenarios, in which facial images from ID documents are compared with digital self-portraits (selfies) for the automated opening of new checking accounts, e.g, or financial transactions authorization. Actually, the comparison of selfies and IDs has also been applied in another wide variety of tasks nowadays, such as automated immigration control. The major difficulty in such process consists in attenuating the differences between the facial images compared given their different domains. In this work, in addition to collecting a large cross-domain face dataset, with 27,002 real facial images of selfies and ID documents (13,501 subjects) captured from the databases of the major public Brazilian bank, we propose a novel architecture for such cross-domain matching problem based on deep features extracted by two well-referenced Convolutional Neural Networks (CNN). Results obtained on the dataset collected, called FaceBank, with accuracy rates higher than 93\%, demonstrate the robustness of the proposed approach to the cross-domain face matching problem and its feasible application in real banking security systems.
\end{abstract}


\IEEEpeerreviewmaketitle

\section{Introduction}
In the last decades, Biometrics emerged as a robust solution for automated people recognition. Among the main biometric traits, face is one of the most convenient since its capture does not require much user collaboration and cameras are present almost everywhere, including in mobile devices \cite{Jain1, Face2, Face3}. Currently, state-of-the-art methods for face recognition and authentication are based on Convolutional Neural Networks (CNN) \cite{lecun, vgg, openface}, deep neural networks inspired on the inner working of human brain, which have presented great accuracy results in many complex tasks involving images. CNNs have been applied in different face recognition and authentication systems, including in commercial ones.

According to \cite{FEBRABAN1}, financial institutions must have effective and reliable methods to authenticate their customers. An effective authentication system should protect customers' data, prevent money laundering and terrorist financing, reduce fraud, inhibit identity theft and promote the legal enforceability of the agreements on electronic transactions \cite{FFIE}. Performing financial transactions with unauthorized or improperly identified people in a banking environment results in huge financial losses, damage to the reputation of the financial company, and breach of bank secrecy. 

In this context, banks are investing in robust methods for automated face-based authentication \cite{Face3} in order to improve the user experience of their systems, especially in mobile banking, as well as to prevent identity theft and fraud. A tendency nowadays in the financial industry is the usage of facial images from different sources (cross-domain problem), usually photographs of ID documents and digital  self-portraits (selfies), for user authentication in order to allow the automated opening of new checking accounts, e.g., or the authorization of financial transactions and the registration of mobile devices~\cite{folego}. Actually, the comparison of selfies  and IDs for people  recognition has also  been applied to  another wide variety of tasks nowadays, such as the automated immigration control, performed by many countries in our days, especially in their airports. 

In this work, in addition to collecting a novel and large cross-domain face database, which we called FaceBank, we propose a novel architecture for the cross-domain face matching problem based on two well-referenced CNNs, VGG-Face \cite{vgg}, and OpenFace~\cite{openface}, by extracting deep and robust features from the facial images, combining them, and by training effective classifiers in order to identify genuine and imposter cross-domain matchings, a complex problem given the significant differences in the facial images captured from the different sources. We also apply some normalization techniques to the images and to their feature vectors in order to better address the cross-domain issues. Results show that the proposed architecture presented great accuracy rates, higher than 93\%, and low processing times, being suitable for the usage in real security systems. Although VGG-Face has been a bit more accurate than OpenFace for feature extraction, the latter one is much more efficient. The FaceBank dataset is composed of 27,002 real facial images of selfies and ID documents (13,501 subjects) captured from the systems of the major public Brazilian bank (the largest dataset of this kind, to the best of our knowledge).

This paper is organized as follows: Section~\ref{fraud} presents an overview on banking systems and identity fraud; Section~\ref{sec:RW} briefly describes some studies on cross-domain face matching; Section~\ref{sec:dataset} presents the FaceBank dataset; Section~\ref{sec:method} describes the proposed approach for cross-domain deep face matching; Sections~\ref{sec:results} and \ref{sec:conclusion} present the experiments, results, some discussions and conclusion.  

\section{Banking Security Systems and Identity Fraud}
\label{fraud}

An increase in occurrences of identity fraud has been observed around the world in the last years. According to the Center for Counter Fraud Studies from the University of Portsmouth, identity fraud has grown steadily over the past 10 years and the estimated damages reached about 5.4 trillion dollars per year, on average, during this period. Since 2009, losses owing to fraud have risen by 56.5\%~\cite{fraudindicator2019}.

Cybercriminals are diversifying their targets and using stealthier methods to commit identity theft and fraud, It is possible to notice an epidemic increase in the number of fraud attempts, with systematic and specialized methods developed by attackers. The 2019 Identity Fraud Study~\cite{identityfraud2019} shows that despite the efforts of the global community, in 2018 about 14.4 million consumers were victims of identity theft or fraud and 23\% of fraud victims had unreimbursed personal expenses. When the subject is about new account fraud (fraudsters open new accounts under victims' names), these losses increased from 3 billion dollars in 2017 to 3.4 billion dollars in 2018 and the most common targets for new account fraud are mortgages, student loans, car loans and credit cards.

A recent report~\cite{experian2020} suggested that 95\% of businesses are confident they can identify and recognise their customers, but only 55\% of consumers don't feel recognized. This report also highlighted that 81\% of consumers view physical biometrics as the more secure form of identity verification.

The increasing costs of electronic fraud to the Brazilian banks, for instance, reach million dollars per year. Given the emergent mobile banking in Brazil, which involves about 31.3 billion transactions~\cite{FEBRABANREPORT}, some Brazilian banks are already using fingerprint authentication in order to achieve a better level of security, as in other countries, in their applications. The customer, previously registered, can carry out financial transactions on mobile devices or even on ATMs by presenting their registered fingers to the sensors. However, not all mobile phones being used in poor countries present fingerprint sensors, despite almost all of them have digital cameras. In the case of ATMs, people need to touch a (not always clean) surface, making customers often unsatisfied with the identification system. Besides, usually ID documents do not present the fingerprints of their owners, making such traits unfeasible for the opening of new checking accounts through mobile devices remotely~\cite{FEBRABAN2}.  

\begin{figure}[h]
    \begin{center}
        \includegraphics[width=0.47\textwidth]{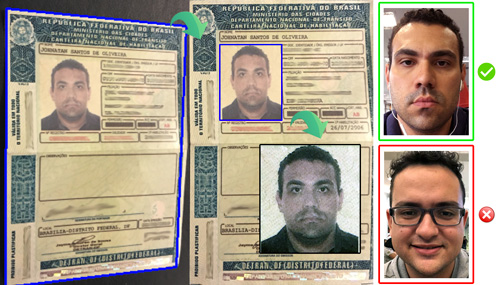}
        \caption{Illustration of the matching process of faces from an ID document and two selfies (from different people). After detecting, cropping and normalizing the face in the document, it is matched with the face of the selfie in order to authenticate the person, validating the document. It is possible to observe different visual aspects in the two kinds of images (ID and selfies) taken with the same mobile phone.}
        \label{fig:process}
    \end{center}
\end{figure}

Due to all these reasons and considering the convenience of face for people recognition and authentication, financial institutions around the world are increasingly implementing tools to allow automated opening of checking accounts, authorization of transactions and devices totally online and through smartphones by means of face authentication. People interested in such services do not need to go to a physical branch to present the required documentation. Instead, using their mobile phones, they can take photographs of their ID documents, containing their facial images, and their digital self-portraits (selfies), proving the possesion of the documents by their legal owners \cite{Face2}. The matching of the faces in the photographs taken can occur directly on the device as well as on the bank server, as shown in  Fig. \ref{fig:process}, authorizing the transactions. Face is a tendency as biometric trait for people authentication in banking environments \cite{Face2}. 

From November of 2016 to January of 2020, the major public Brazilian bank received about 4.1 million requests for opening checking accounts through smartphones, 1.1 million only in 2019 (50\% more than 2018). All of them being manually inspected until now. Among this total, 81\% of these new customers are under 40 years old and 22\% of them were rejected by the human experts. Besides presenting different faces in the ID document and selfie (indicating fraud), most of the requests presented low quality facial images due to issues with illumination, facial occlusion, expression changes, low resolution, or even due to scratched documents. Since the matching is still performed by humans and given the high number of requests being received, such a process is quite expensive for the bank, slow and also subject to failures. An automated method could, at least, automatically discard some of the requests, saving time and resources for the financial institution. By the end of 2020, the bank expected to to reach a total of 6.5 million opened checking accounts through mobile devices.

\section{Cross-Domain Face Matching}
\label{sec:RW}

As previously stated, face is one of the most convenient biometric traits since its capture can be performed at a distance, in a non-intrusive and even non-cooperative way \cite{Jain1, Face2}. Besides,  cameras are nowadays found almost everywhere, including in mobile devices, the main technological basis for the present and future of banking transactions \cite{FEBRABAN1}. When the work involves comparing cross-domain images (e.g., matching faces obtained from ID documents and selfies, or faces extracted from surveillance videos), it is possible to note a substantial increment on the difficulty of the matching process, which is already a complex problem by itself. In cross-domain conditions, the classification algorithms usually have their performance deteriorated due to the different visual aspects of the facial images from different sources, such as different kind of blur, illumination changes, noise or even change on the facial expressions.

Folego et al. \cite{folego} explored approaches for cross-domain face authentication, comparing selfies to ID photographs based on features extracted by the VGG-Face \cite{vgg} deep neural network. They approach the problem with proper image photometric adjustments and data normalization techniques, together with such deep learning architecture, to extract the most prominent and robust features from the original images, reducing the effects of domain differences. However, their dataset was composed of relatively few images (dozens of individuals) and not obtained from a real scenario.

In another work on cross-domain face matching, in order to deal with typical face cross-domain issues such as illumination, alignment, noise, or even facial expression changes, Ho and Gopalan \cite{gopalan} propose to derive a latent subspace for the original faces, characterizing their multifactor variations. Images were synthesized by the authors in order to produce different illumination and other 2D perturbations, forming tensors to represent the faces. Results indicated that the method is effective on constrained and unconstrained datasets.
 
To the best of our knowledge, no evaluation regarding cross-domain face matching on large datasets with real facial images (real selfies and ID documents) was performed and reported in literature, one of the main contributions of our work.

\section{FaceBank Dataset}\label{sec:dataset}

The amount of data for training is an important issue when dealing with Machine Learning algorithms, especially with deep learning-based approaches. Given the high capacity of the deep neural networks due to their large amounts of free parameters to be tuned, the quality of their predictions improves with experience \cite{Kriz}. Face recognition and authentication systems built by large private corporations present, in general, top accuracies, since they are trained on huge private datasets, containing millions of facial images, usually obtained from social media, far more than the number of images in the datasets usually available for research.

Regarding cross-domain face authentication (comparison of ID document photographs and selfies), there is no large dataset available for tests either. Usually, given the difficulty to collect data, researchers evaluate their new methods with images from few individuals or with synthetic datasets generated by applying a set of mathematical transformations on a small subset of real images. No large dataset with real images from banking systems was also reported in literature in the past.

Based on these considerations, we obtained authorization from the largest public Brazilian bank to collect a large cross-domain dataset, which we called FaceBank, from its databases of facial images (selfies and scanned ID documents), in order to conduct this work. Initially, about 150,000 images in RGB color space were collected among selfies, from the profiles of individuals in the bank internal social network, and photographs of the ID documents of the same individuals. However, we detected that many of these images presented no faces (especially the profile images) or faces with low resolution (in the case of the IDs). In this sense, in order to eliminate such bad images and avoid a decrease in the performance of our proposed architecture, and also given the processing and time restrictions for collecting the images we had to follow, we applied a fast technique based on the efficient face detection algorithm of Viola and Jones~\cite{violajones} to the original set of images in order to detect which of them presented real faces with, at least, regular resolution. We discarded both images of the users that had at least one image (selfie or ID) discarded by such method. 

\begin{figure}[h]
    \begin{center}
        \includegraphics[width=0.48\textwidth]{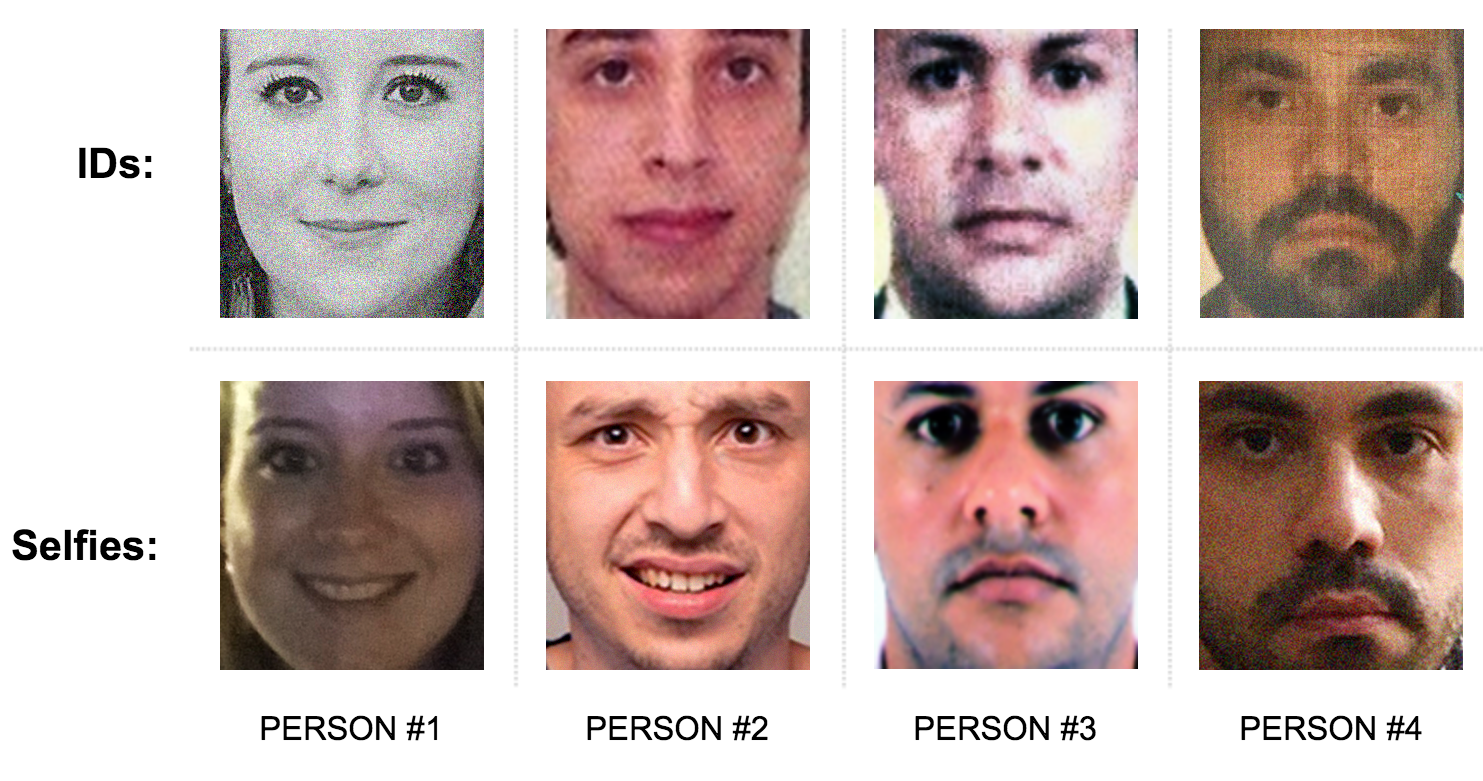}
        \caption{Examples of real facial images from selfies and ID documents of the FaceBank dataset. The dataset contains a total of 27,002 images (13,501 individuals).}
        \label{fig:datasetBB}
    \end{center}
\end{figure}

After this process, we obtained 27,002 images from 13,501 subjects (two images per subject, i.e., selfie and ID document). In order to crop the faces in these remaining images precisely, a more robust algorithm based on HOG (Histogram of Oriented Gradients) \cite{HOG} was applied to detect the faces again in such images as well as some landmark points (such as eyes and mouth coordinates, etc.). The cropped faces composed the FaceBank dataset and some of them are shown in Fig. \ref{fig:datasetBB}. Even visually, it is possible to note the huge differences in the facial images of the same person from different domains, i.e., selfie and ID document, and also the regular quality of the resultant images, typical from real banking scenarios, all this demonstrating the high complexity of the cross-domain face authentication problem. 

\section{Proposed Approach}\label{sec:method}

In this work, besides of collecting the large FaceBank dataset with real banking facial images of selfies and ID documents, we also propose a robust architecture for cross-domain face matching based on two well-referenced Convolutional Neural Networks (CNN), VGG-Face \cite{vgg} and OpenFace \cite{openface}, to extract deep and robust features from the faces, with good level of invariance to the domain differences. We also applied normalization techniques to the facial images and to their feature vectors in order to attenuate such issues even more and also improve the model performance. 

After normalizing the selfies and IDs, extracting and normalizing their deep feature vectors using the VGG-Face \cite{vgg} or OpenFace \cite{openface} CNN models, we trained and assessed four classifiers (Linear Support Vector Machine - Linear SVM \cite{svm}, Power Mean SVM - PmSVM \cite{pmsvm}, Random Forest - RF \cite{RF}, and RF with Ensemble Vote Classifier - Voting RF \cite{raschka}) in order to verify which one performed better in the task of classifying a pair of face images (selfie and ID) as genuine or impostor and compare their results.

\begin{figure*}[h]
    \begin{center}
        \includegraphics[width=0.98\textwidth]{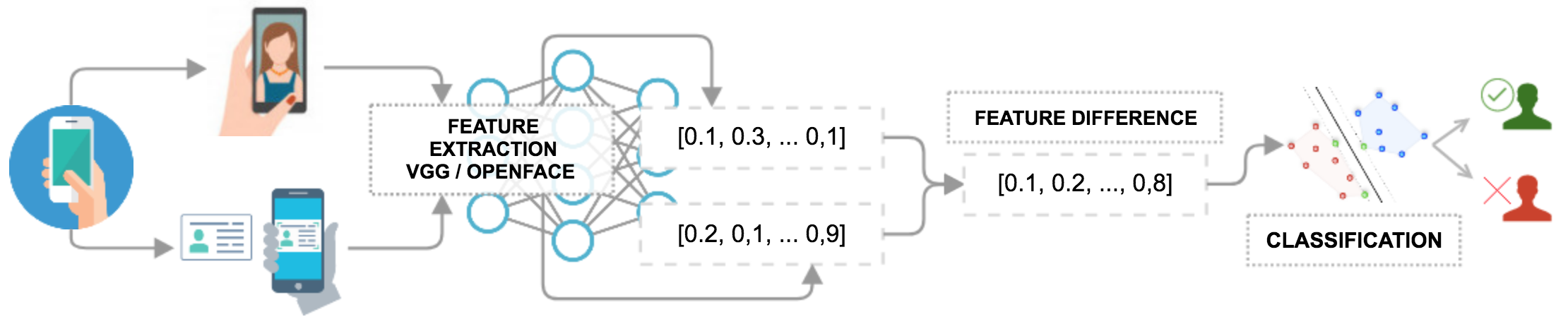}
        \caption{Overview of the proposed architecture for cross-domain face matching: facial images are taken with the mobile phone and normalized; their features vectors are extracted using deep neural networks (VGG-Face or OpenFace); they are also normalized and subtracted from each other; and then the final difference feature vector is classified as a genuine access (same person in both images) or impostor request (distinct people in the images) by a classifier.}
        \label{fig:pipeline}
    \end{center}
\end{figure*}

In summary, given a test pair of selfie and ID document images, a sequence of steps including face normalization, deep features extraction, feature vectors normalization, as well as classification, are performed in order to verify whether such pair of facial images were captured from the same person (genuine pair) or from different individuals (impostor pair). Fig. \ref{fig:pipeline} shows these steps (proposed architecture), which are described in subsections \ref{A} to \ref{E}.

\subsection{Face Detection, Cropping and Alignment}\label{A}

In order to carry out face detection and cropping, as said, in this work we used a robust and efficient algorithm available in the Dlib library \cite{dlib}, which is based on HOG (Histogram of Oriented Gradients) \cite{HOG} features. This algorithm returns the coordinates of the rectangle that contains the detected face in the input image, as well as the coordinates of the left and right eyes. With this information it is possible to align, crop and resize all the face images from the dataset before starting to compare them.

Likewise in \cite{folego}, in the face cropping step, we included the ear, chin, and hair in the Region of Interest (ROI), by expanding by 22\% the initial rectangle returned by the algorithm of the Dlib. This expanded ROI tends to increase the results of the face matching. 

Face alignment is performed by rotating the face until the coordinates of both eyes are in line with the $x$-axis. Finally, the cropped and aligned face images are resized by using bilinear interpolation. For the VGG-Face feature extraction based approach, the face image size must be $224\times224$ pixels, whilst for the OpenFace feature extraction based approach the face image size must be $96\times96$ pixels.

\subsection{Face Normalization}\label{B}

A typical problem of comparing photos of documents with selfies is the large difference in lighting due to the change in the source domain. Other issues, such as facial pose and expression changes, as well as the different resolutions of the images are also problematic. Aiming to mitigate some of these problems, especially illumination differences, and before extracting the features of the faces in the images, we applied the Automatic Color Equalization (ACE) \cite{ace} to normalize the cropped facial images from the selfies and IDs. 

The ACE technique is based on a computational model of the human visual system that performs a photometric transformation on the images in order to equalize simultaneously global and local effects of illumination \cite{ace}. It obtains good contrast enhancements even when the quality of the images is poor. We use this technique as an effort to approximate the two kinds of images under analysis.

\subsection{Data Augmentation}\label{C}

Obtaining a huge amount of real data (of real training samples) is an expensive and not always a possible process. Aiming to add even more data to the 27,002 images of the FaceBank dataset, we generated new ones by applying some transformations on them. This data augmentation strategy is  a  common practice when working with CNNs and other classifiers \cite{AugmentationCNN}.

According to Nielsen~\cite{nielsen}, despite the fact that artificial images do not substitute the potential of real samples, it is conceivable that adding, to the training data, transformed images based on the original ones might help the deep neural networks learn more about the patterns being addressed. By making small modifications on the original images, it is possible to expand the training database substantially. Common augmentation methods include noise addition, image equalization, random crop, scale change, jitter, brightness and contrast modifications. In this work, three of these methods were applied. 

Initially, we increased the FaceBank dataset by adding white Gaussian noise to the original facial images in the regions near the eyes. As a generic approach, we used a sampling mechanism that added uncorrelated Gaussian noise \(\alpha\) to the visual input $x$. If $k$ indexes the raw pixels, a new sample is given by: $x_k = x_k + \alpha_k$.

The second transformation, applied to the original and noisy images, was to randomly increase or decrease their brightness, so that the model would learn not to rely on brightness information. As in \cite{du_guo}, new images were generated with different brightness by first converting the images to the HSV (Hue, Saturation, and Value) color space and scaling the V channel up or down (converting the image back to the RGB color space after that).

Finally, in order to further augment the database obtained after the two previous transformations, we applied the Contrast Limited Adaptive Histogram Equalization (CLAHE) \cite{clahe} technique to the training images, which divides the input image into small blocks, applies a conventional histogram equalization in each block, and then checks if any histogram \textit{bin} is above the contrast limit. At the end of the data augmentation process, we obtained $216,016$ images ($27,002\times2^3$, since we applied $3$ transformations doubling the size of the dataset in each of them), $108,008$ selfies and $108,008$ ID documents. This set of images was called Augmented FaceBank.

\subsection{Feature Extraction, Normalization and Difference}\label{D}

In order to extract robust features from the facial images given their different domains, we used the well-referenced Convolution Neural Network (CNN) called VGG-Face \cite{vgg}, originally trained using a dataset with more than 2.6 million (same domain) facial images of 2,622 different people, which achieved state-of-the-art results in face recognition. By using the trained VGG-Face model, a very deep model of CNN containing 16 layers, we avoided many issues such as overfitting in our dataset, despite its size, as well as obtaining a good power of generalization due to the high capacity of the network (huge amount of parameters) and its large original training set. 

We used the trained model of VGG-Face for Transfer Learning, i.e., we passed our facial images (from the Augmented FaceBank dataset) through the network and extracted their feature vectors based on the output of the layer ``fc6" of the network (the third layer from top to bottom). Despite the fact that other studies usually extract features from the layer ``fc7" (on top of ``fc6") of the VGG-Face trained model when performing such task, we explored the layer ``fc6", a fully connected layer with 4,096 neurons, since in \cite{folego} it allowed obtaining the best results for cross-domain face matching.

In order to compare the results with the performance of a different deep neural network, in this work we also assessed another well-referenced CNN: OpenFace \cite{openface}. OpenFace is an open source model, also implemented and trained on datasets of facial images from the literature. Besides being able to use this neural network in commercial applications due to its open license, another interesting aspect of OpenFace is that it maps each face into an Euclidean space (into a hypersphere within it) by a $128$-dimensional feature vector, output of its top layer. Its training algorithm, is mainly based on the Triplet Learning \cite{facenet} approach, in which the network is trained on genuine (same person) and imposter (different people) pairs of faces and tries to ensure that the faces of genuine pairs are closer in such Euclidean space than faces from different people, given a tolerance margin, following Eq. \ref{triplearning}:

\begin{equation}
\label{triplearning}
\|x^a-x^p\|_2+\alpha < \|x^a-x^n\|_2,  \forall x^a, x^p, x^n \in \tau,
\end{equation}
where $x^a$ and $x^p$ indicate feature vectors of faces from the same person, $x^n$ the feature vector from the face of another person, $\alpha$ is the tolerance margin (usually set to $0.2$), and $\tau$ the training set for the neural network. The similarity degree of two faces is measured based on the Euclidean distance between their feature vectors.

The VGG-Face model presents more parameters (higher capacity for feature learning) and allows extracting larger feature vectors. However, OpenFace (its default model), besides presenting a slightly different training algorithm and open license, is also interesting to our problem due to its great results reported in other applications \cite{openface} and efficiency, being especially suitable for mobile banking.   

Given the cross-domain problem, feature vectors extracted from images of different domains might have values with significantly different magnitudes. To mitigate this problem when comparing such vectors, we applied $L_2$ normalization to them. The \(p\)-norm of a feature vector \(x  \in  \Re^n\) is given by:
\begin{equation}
\vert\vert x \vert\vert_p = (\sum_{i=1}^{n}  \vert x_i \vert^p)^{\frac{1}{p}},    
\end{equation}
where $n=4,096$ for VGG-Face and $n=128$ for OpenFace. 

The $L_2$-normalized version of each feature vector $x$ is given by:
\begin{equation}
\hat{x} = \frac{x}{\vert\vert x \vert\vert_2}. 
\end{equation}

After normalizing the feature vectors of the faces from the Augmented FaceBank, for each pair of selfie and ID, we also combined their vectors, \(a\) and \(b\), respectively, into a final feature vector in order to emphasize their different properties and train the classifiers (in order to identify genuine and impostor face matchings), by using the absolute value of the subtraction \(f_{ab} = \vert a - b \vert\), which showed to be the best choice for our problem.

\subsubsection{Pair Generation}
\label{pair}

As said, to train the classifiers, we extracted the deep features of each pair of faces (selfie and ID) from the Augmented FaceBank using one of the CNNs evaluated. Then, the extracted features were normalized and stored into feature vectors. In order to verify whether a pair of selfie and ID images is from the same person (genuine matching), or from distinct people (imposter matching), the ID feature vector is subtracted from the selfie feature vector and the vector resultant from the module of the difference is finally presented to the classifier. For the pair generation task, we performed a random split of the individuals of the dataset into two disjoint sets: training and test. The training set contained 80\% of the individuals of the Augmented FaceBank dataset, while the test set had 20\%. Then, in each set, we generated random pairs of two images (selfie and ID) for representing genuine matchings and imposter matchings.

\subsection{Classification}\label{E} 

Given the genuine and imposter pairs of selfies and IDs and their difference vectors, we assessed different and well-referenced classifiers to determine the best option for our cross-domain face matching problem. We selected four effective and also efficient classifiers from the literature in order to evaluate their performances in our cross-domain problem: Linear Support Vector Machine (Linear SVM) \cite{svm}; Power Mean SVM (PmSVM) \cite{pmsvm}; Random Forest (RF) \cite{RF}; and RF with Ensemble Vote Classifier (Voting RF) \cite{raschka}. In our case, the Voting RF combines the decisions of 5 RFs. Due to robustness and efficiency of the code and reprodutibility of the experiments, we used the implementations of such methods available in the well-referenced Scikit Learning library \cite{sk}. 

Likewise in \cite{folego}, we also decided to evaluate the Linear and PmSVM given their good performances in many tasks and due to their reported efficiency. Despite that the Linear SVM presents inferior results in many tasks than SVMs with other kernels, it is fast, being more appropriate for environments with hardware restrictions, as in mobile devices. Regarding PmSVM, compared with state-of-the-art methods for large-scale image classification, it has achieved the highest learning speed and highest accuracy in many cases \cite{jianxin}. The RF-based classifiers are also robust and very efficient since they are based on decision trees. 
 
In order to measure the accuracy of the selected classifiers, we used the global accuracy metric since some of them only had as output the class of each test sample, following Eq. \ref{eqAcc}:

\begin{equation}
\label{eqAcc}
accuracy = \frac{1}{n_{test}}\sum_{i=1}^{n_{test}} 1(\hat{y}_i = y_i),
\end{equation}
where \(\hat{y}_i\) is the predicted label for the $i^{th}$ test sample, \(y_i\) is the real label of such sample and $n_{test}$ is the number of test samples.

\section{Experiments, Results and Discussion} \label{sec:results}

In order to assess the proposed approach and the performance of the assessed classifiers and to analyze the feasibility of its application in real banking security systems, especially for mobile devices, we considered one imposter pair for each genuine pair in the training and test stages of the classification, for a balanced training.

We evaluated the performance of the proposed architecture more than once by varying the number of subjects and the total number of difference feature vectors being considered, in order to verify, in a more detailed way, their robustness regarding the amount of data for training and test. For all classifiers, we used the default hyper-parameters defined on the Scikit Learning library \cite{sk}. For the PmSVM, the default value of the regularization parameter \(\omega\) was set to 0.01. Tab. \ref{tb_1p1n} shows the results obtained for all the classifiers given the features extracted by VGG-Face. In the first test, for example, we considered only 10,000 subjects from the Augmented FaceBank dataset (20,000 pairs of faces, 10,000 genuines and 10,000 imposters). We set, as said, 80\% of the subjects (and their respective difference feature vectors) for training and 20\% for test.

\begin{table}[h]
\centering
\caption{Accuracy results (\%), given the features extracted by VGG-Face, on the Augmented FaceBank dataset, considering one imposter pair for each genuine pair of faces (selfie and ID) and varying the number of subjects and difference feature vectors under analysis. The best result for each classifier is highlighted.}
\label{tb_1p1n}
\resizebox{0.47\textwidth}{!}{%
\begin{tabular}{|c|c|c|c|c|c|}
\hline
\textbf{ Subjects} & \textbf{\begin{tabular}[c]{@{}c@{}} Pairs\\ Train/Test\end{tabular}} & \textbf{\begin{tabular}[c]{@{}c@{}}Linear \\ SVM\end{tabular}} & \textbf{PmSVM} & \textbf{RF} & \textbf{Voting RF} \\ \hline\hline
$10,000$           & \begin{tabular}[c]{@{}c@{}}$16,000$ /\\ $4,000$\end{tabular} & 91.65 & 89.57 & 89.65 & 93.45  \\ \hline
$20,000$   & \begin{tabular}[c]{@{}c@{}}$32,000$ /\\ $8,000$\end{tabular} & 92.43 & 89.87 & 89.13 & 93.28            \\ \hline
$50,000$   & \begin{tabular}[c]{@{}c@{}}$80,000$ /\\ $20,000$\end{tabular} & 92.69 & 88.09 & 89.26 & \textbf{93.51}
    \\ \hline
$80,000$   & \begin{tabular}[c]{@{}c@{}}$128,000$ /\\ $32,000$\end{tabular} & 92.75 & 89.87 & \textbf{89.77} & 92.67           \\ \hline
$108,008$          & \begin{tabular}[c]{@{}c@{}}$172,808$ /\\ $43,208$\end{tabular} & \textbf{92.81} & \textbf{90.91} & 88.95 & 92.82
    \\ \hline
\end{tabular}}
\end{table}

As one can observe, the Voting RF obtained the best overall performance and its best accuracy result occurred when we considered 50,000 subjects. When working with 108,008 individuals, this classifier presented only a slight decrease in performance compared with the previous tests, still presenting better results than all other classifiers and demonstrating its robustness. Regarding processing time, voting RF is also very efficient by working with decision trees. It spent, on average, only 30 milliseconds for classification of each test sample. 

As shown, the performance of the Linear SVM and PmSVM, in general, increased with the increasing sizes of the training and test sets, also demonstrating their robustness to large datasets (often found in real scenarios), despite being slower the former classifier (Linear SVM spent about 45 milliseconds for each test samples). The RF classifier presented its best performance with 80,000 subjects.

The results obtained by the classifiers given the feature vectors extracted by OpenFace are shown in Tab. \ref{tb_1p1n_openface}. It is important to note that the results obtained were very close to those of VGG-Face. Besides, OpenFace works by default with smaller images ($96\times96$ pixels), saving computational resources in the forward pass of the images through the network for feature extraction, and generates a much more efficient representation for the faces (it generates a 128-dimensional feature vector for each face, and VGG-Face a 4,096-dimensional vector), allowing classifiers being faster and more suitable for mobile applications. The forward pass of each facial image in the VGG-Face took about 2.89 seconds while in OpenFace it took only 0.14 seconds per image.

\begin{table}[h]
\centering
\caption{Accuracy results (\%), given the features extracted by OpenFace, on the Augmented FaceBank dataset, considering one imposter pair for each genuine pair of faces (selfie and ID) and varying the number of subjects and difference feature vectors under analysis. The best result for each classifier is highlighted.}
\label{tb_1p1n_openface}
\resizebox{0.47\textwidth}{!}{%
\begin{tabular}{|c|c|c|c|c|c|}
\hline
\textbf{ Subjects} & \textbf{
\begin{tabular}[c]{@{}c@{}} Pairs\\ Train/Test\end{tabular}} & \textbf{\begin{tabular}[c]{@{}c@{}}Linear \\ SVM\end{tabular}} & \textbf{PmSVM} & \textbf{RF} & \textbf{Voting RF} \\ \hline\hline
$10,000$           & \begin{tabular}[c]{@{}c@{}}$16,000$ /\\ $4,000$\end{tabular} & 89.17 & 85.45 & 88.72 & \textbf{91.50}            \\ \hline
$20,000$   & \begin{tabular}[c]{@{}c@{}}$32,000$ /\\ $8,000$\end{tabular} & \textbf{89.91} & 85.67 & 89.48 & 90.65            \\ \hline
$50,000$   & \begin{tabular}[c]{@{}c@{}}$80,000$ /\\ $20,000$\end{tabular} & 89.82 & 86.81 & 89.17 & 90.86
    \\ \hline
$80,000$   & \begin{tabular}[c]{@{}c@{}}$128,000$ /\\ $32,000$\end{tabular} & 89.88 & \textbf{86.89} & \textbf{89.52} & 90.70           \\ \hline
$108,008$          & \begin{tabular}[c]{@{}c@{}}$172,808$ /\\ $43,208$\end{tabular} & 89.89 & 86.73 & 89.39 & 90.61
    \\ \hline
\end{tabular} }
\end{table}

The best result regarding all experiments, $93.51\%$ of accuracy, was obtained by the VGG-Face neural network with the Voting RF classifier when working with $50,000$ subjects. Voting RF obtained the best results in all experiments, with both CNNs, being very suitable for the cross-domain face matching problem due to its efficiency inherited from the decision trees.

\begin{figure}[b]
    \begin{center}
        \includegraphics[width=0.475\textwidth]{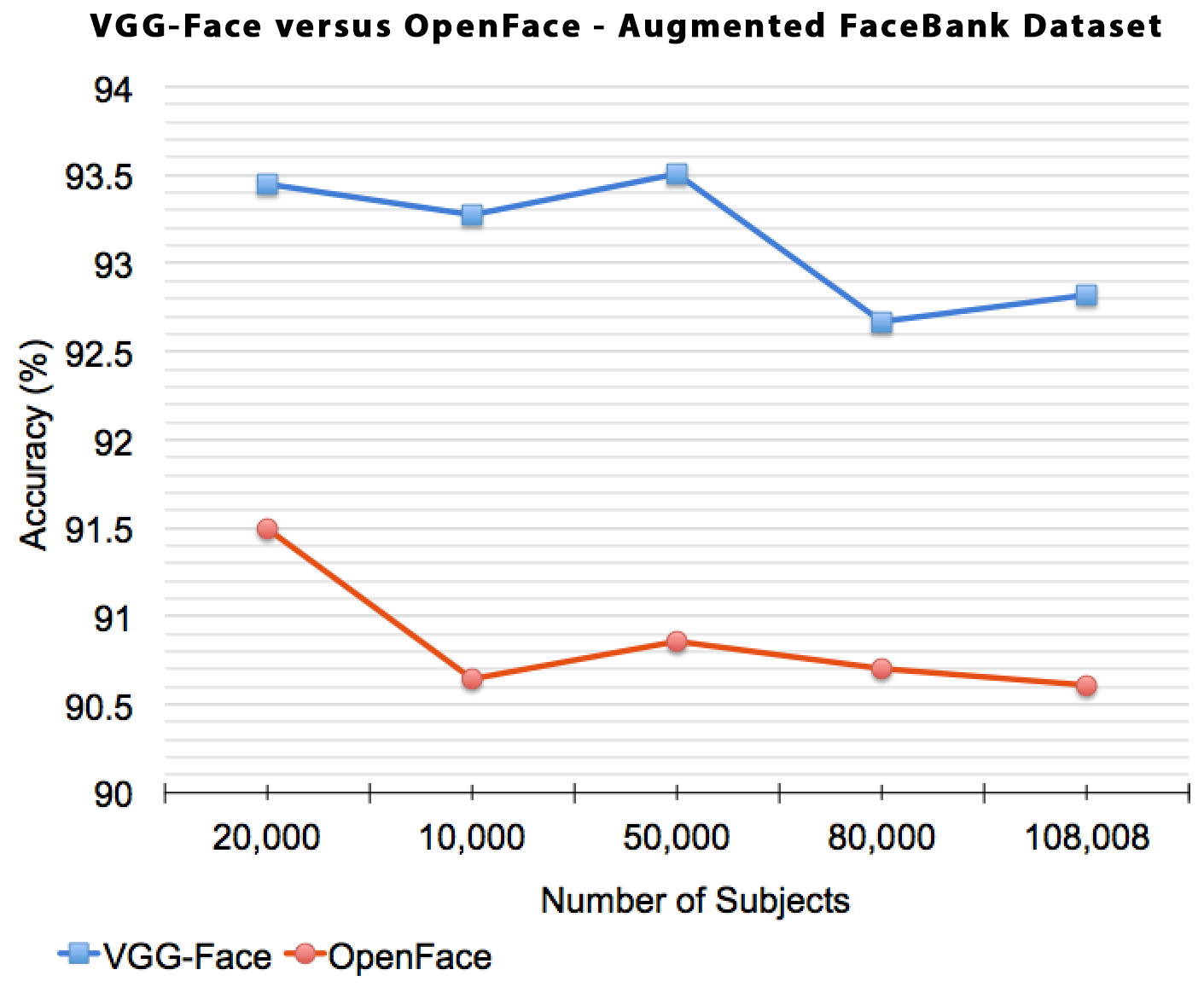}
        \caption{Comparison of the performance (global accuracy) of the Voting RF classifier given the feature vectors (their difference versions) extracted by each CNN: VGG-Face and OpenFace.}
        \label{fig:grafico}
    \end{center}
\end{figure}

In order to better visualize the performances of both evaluated CNNs with such powerful classifier (Voting RF), Fig. \ref{fig:grafico} shows the accuracies obtained by this classification method by varying the size of the dataset. As can be seen in Fig \ref{fig:grafico}, the Voting RF classifier tends to decrease its performance, as expected, when considering more subjects. However, such deterioration in accuracy is not so accentuated for both CNNs.  

\begin{figure}[!ht]
    \center
    \includegraphics[width=0.55\textwidth]{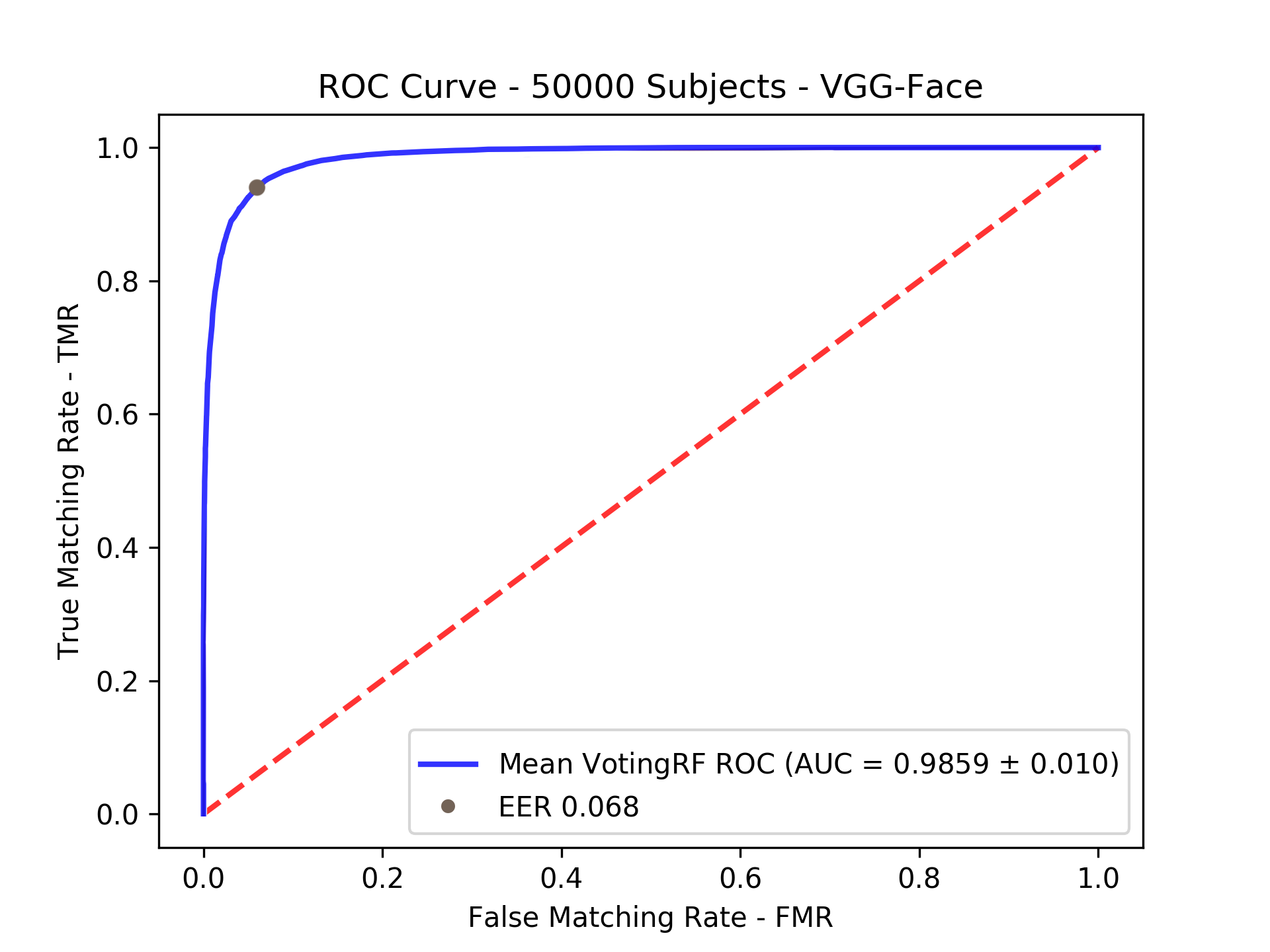}
    \caption{Receiver Operating Characteristics (ROC) curve for the Voting RF classifier given the features extracted by VGG-Face on the Augmented FaceBank dataset, considering one negative pair for each positive pair, with 50,000 subjects and 100,000 combined vectors (total number of pairs) under analysis.}\label{fig:rocVGG}
\end{figure}

\begin{figure}[!ht]
    \center
    \includegraphics[width=0.55\textwidth]{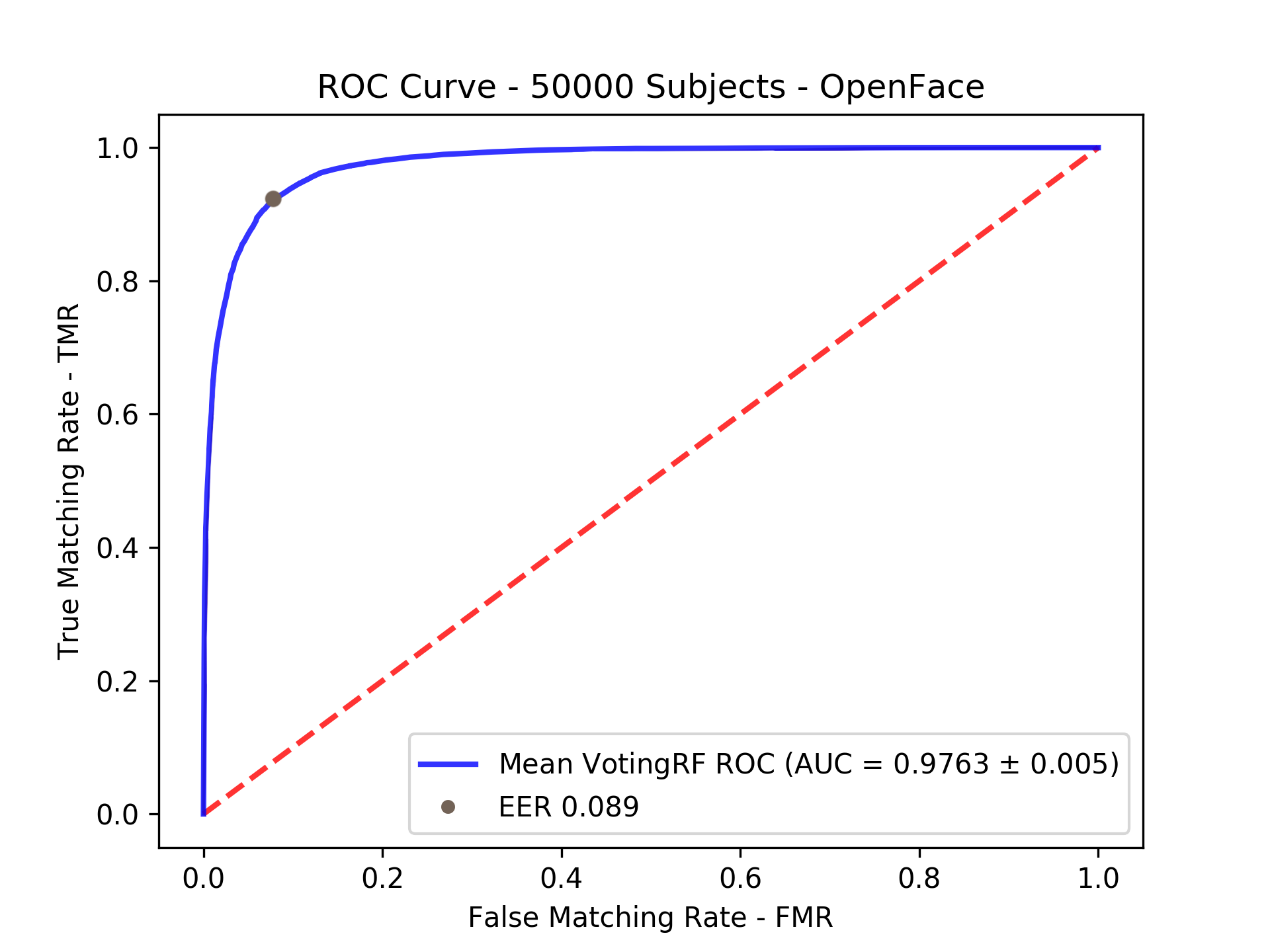}
    \caption{Receiver Operating Characteristics (ROC) curve for the Voting RF classifier given the features extracted by OpenFace on the Augmented FaceBank dataset, considering one negative pair for each positive pair, with 50,000 subjects and 100,000 combined vectors (total number of pairs) under analysis.}\label{fig:rocOpenface}
\end{figure}

For a more detailed analysis of the performance of the Voting RF classifier (best classification method in the experiments) working with both CNNs, VGG-Face and OpenFace, on the balanced dataset presented, in terms of True Matching Rate (TMR), i.e., probability of correct match, and False Matching Rate (FMR), probability of incorrect match, and verify the effect of varying the decision (acceptance) threshold of the system, we also built its Receiver Operating Characteristics (ROC) \cite{metz1978basic} curves, which are shown in  in Fig. \ref{fig:rocVGG} (working with the VGG-Face neural network) and Fig. \ref{fig:rocOpenface} (working with OpenFace). The Equal Error Rate (EER) of each curve, which indicates the error rate when the False Non-Matching Rate (FNMR), i.e., FNMR=1-TMR, and the FMR of the system are the same, is also reported in such images.

\section{Conclusion} \label{sec:conclusion}

Banking identity fraud has become increasingly common worldwide, causing huge financial losses to the banks and financial systems, making them invest massively in higher-level security systems, mainly based on biometric recognition. Among the main traits, face is one of the most important due to its convenience and the  availability of digital cameras almost everywhere, including in mobile devices. Besides, a tendency nowadays is to open new checking accounts through smartphones in an automated way by matching facial images from selfies and photographs of ID documents. Such cross-domain problem is a highly complex task due to differences between the two kinds of images. 

In this work, we collected a large dataset, which we called FaceBank, with 27,002 real images among selfies and ID documents (13,501 subjects) from the databases of the largest public Brazilian bank, and proposed a robust architecture for cross-domain face matching, comparing selfies and IDs, based on two well-referenced CNNs, VGG-Face and OpenFace, which obtained great results (accuracy rate higher than 93\%), even in such difficult task. In the LFW benchmark, OpenFace has achieved about 92\%, while VGG-Face 98\%. To the best of our knowledge, FaceBank is the largest cross-domain face dataset collected, with real banking images, and this is the first large scale study on such kind of dataset. We plan to make the collected database available for the use of other researchers and the scientific community.

The usage of deep facial features extracted by well-referenced CNNs, VGG-Face and OpenFace, proper image processing techniques, feature vectors normalization and their differences calculation (by pairs of images), robust classifiers, especially the Voting RF, attenuates significantly the effects of domain differences, allowing  good results even when working with a large number of facial images. Based on the accuracy obtained (higher than 93\%) and its efficiency, it is possible to conclude that the proposed architecture for cross-domain deep face matching is feasible for real applications, especially banking ones. The proposed approach can also be applied, for instance, to help human experts in extremely critical scenarios, rejecting the matchings with very low scores. All this would save crucial time and resources for the financial institutions.


\bibliographystyle{IEEEtran}

\end{document}